\documentclass[11pt,letterpaper]{article}
\usepackage{ijcnlp2017}

\usepackage{times}
\usepackage{latexsym}
\usepackage{graphicx}
\usepackage{caption}
\usepackage{adjustbox}
\usepackage{amssymb}
\usepackage{multirow}
\usepackage{footnote}
\makesavenoteenv{tabular}
\usepackage{subcaption}

\ijcnlpfinalcopy

\DeclareGraphicsExtensions{.pdf,.png,.jpg}

\newcommand\RN{ReasoNet++}

\title{An Empirical Analysis of Multiple-Turn Reasoning Strategies \\ in Reading Comprehension Tasks}

\author{Yelong Shen$^\bold{\dagger}$, Xiaodong Liu$^\bold{\dagger}$, Kevin Duh$^\bold{\ddagger}$, Jianfeng Gao$^\bold{\dagger}$ \\
  $^\bold{\dagger}$    
  Microsoft Research, Redmond, WA, USA \\
  $^\bold{\ddagger}$
  Johns Hopkins University, Baltimore, MD, USA \\
  {\tt $^\bold{\dagger}$\{yeshen,xiaodl,jfgao\}@microsoft.com
   $^\bold{\ddagger}$kevinduh@cs.jhu.edu}
}

\begin{document}

\maketitle

\begin{abstract}
Reading comprehension (RC) is a challenging task that requires synthesis of information across sentences and multiple turns of reasoning. 
Using a state-of-the-art RC model, we empirically investigate the performance of single-turn and multiple-turn reasoning on the SQuAD and MS MARCO datasets.
The RC model is an end-to-end neural network with iterative attention, and uses reinforcement learning to dynamically control the number of turns. 
We find that multiple-turn reasoning outperforms single-turn reasoning for all question and answer types; further, we observe that enabling a flexible number of turns generally improves upon a fixed multiple-turn strategy. 
We achieve results competitive to the state-of-the-art on these two datasets.
\end{abstract}

\section{Introduction}
\label{sec:intro}

There is an old Chinese proverb that says: \textit{``Read a hundred times and the meaning will appear.''}
Several recent reading comprehension (RC) models have embraced this kind of multiple-turn strategy;
they generate predictions by making multiple passes through the same text and integrating intermediate information in the process \cite{hill2015goldilocks,dhingra2016gated,sordoni2016iterative,shen2016reasonet}.
While state-of-the-art results have been achieved by these models, there has yet to be an in-depth analysis of the impact of the multiple-turn strategy to reasoning. 
This paper attempts to fill this gap.

We provide empirical results and analysis on two challenging RC datasets: the Stanford Question Answering Dataset (SQuAD) \cite{rajpurkar2016squad}, and the Microsoft Machine Reading Comprehension Dataset (MS MARCO) \cite{nguyen2016ms}. 
Given a question $Q$, the RC model is to read passages $P$ and produce an answer $A$, which could be free-form text or one of the possible candidate spans in the passage.

The following example from SQuAD illustrates the need for synthesis of information across sentences and multiple turns of reasoning:\\[0.2cm]
$Q$: What collection does \textbf{the V\&A Theator \& Performance galleries} hold?\\[0.2cm]
$P$: \textbf{The V\&A Theator \& Performance galleries} opened in March 2009. ... \textbf{They} hold the UK's biggest national collection of \underline{material about live performance.}\\[0.2cm]
To infer the answer (the underlined portion of the passage $P$), the model needs to first perform coreference resolution so that it knows ``\textbf{They}'' refers ``\textbf{V\&A Theator}'', then extract the subspan in the direct object corresponding to the answer. This process can be modeled by the repeated processing of intermediate states and input in a neural net.

To perform the analysis, we adopt the ReasoNet model of \newcite{shen2016reasonet}. 
This is an end-to-end neural network that uses an iterative attention mechanism to simulate multiple-turn reasoning in RC. 
It has achieved strong results on cloze-style RC tasks like CNN/DailyMail \cite{hermann2015teaching} and we extend it to SQuAD and MS MARCO tasks.
The advantage of using ReasoNet for our purpose is that it uses reinforcement learning to dynamically determine the number of turns for each question-passage pair. This enables us to analyze the behavior of multiple-turn reasoning in neural network models. 

We find that multiple-turn reasoning outperforms single-turn reasoning across the board for various types of question and answer types. Furthermore, the flexibility to dynamically decide the number of turns generally improves over a fixed multiple-turn strategy, where the number of turns are set a priori. 
As an additional contribution, our extension to the ReasoNet model achieves results competitive with the state-of-the-art on SQuAD and MS MARCO.

In the following, Section \ref{sec:task} describes our two RC tasks, Section \ref{sec:model} explains the model we used for analysis, and Section \ref{sec:experiment} discusses results.


\section{Reading Comprehension Tasks}
\label{sec:task}

\begin{table}[t!]
\centering
\begin{tabular}{|@{\hskip2pt} l @{\hskip2pt}||@{\hskip2pt} c @{\hskip2pt}|@{\hskip2pt} c @{\hskip2pt}|}
\hline
                     &     SQuAD          & MS MARCO \\ \hline
query source & crowdsourced & user logs \\
answer ($A$)          & span of words & free-form text \\
\#questions ($Q$)      & 100K questions  & 100K queries \\
\#passages ($P$)  & 23K paragraphs  & 1M paragraphs \\\hline
\end{tabular}
\caption{\label{tab:data} Dataset characteristics} 
\end{table}

We focus this study on two RC tasks which we believe require sophisticated reasoning. \\[0.2cm]
\textbf{SQuAD}: SQuAD  is a machine comprehension dataset constructed on 536 Wikipedia articles (23K paragraphs), with more than 100,000 questions. In contrast to prior datasets such as \cite{richardson13mctest,hermann2015teaching}, SQuAD does not provide a multiple choice list of answer candidates. Instead, the RC model must select the answer from all possible spans in the passage. Crowdsourced workers are asked to read each passage (a paragraph), come up with questions, and then mark the answer spans. 

There is a variety of questions and answers. The authors of SQuAD described several types of reasoning required to answer questions: (a) lexical variation between question ($Q$) and answer ($A$) that can be solved by understanding synonyms, (b) lexical variation that could be solved by world knowledge, (c) syntactic variation between Q and A sentence, and (d) multiple sentence reasoning that require anaphora or higher-level fusion. 



\begin{figure*}[h!]
\centering
  {\includegraphics[scale=0.55]{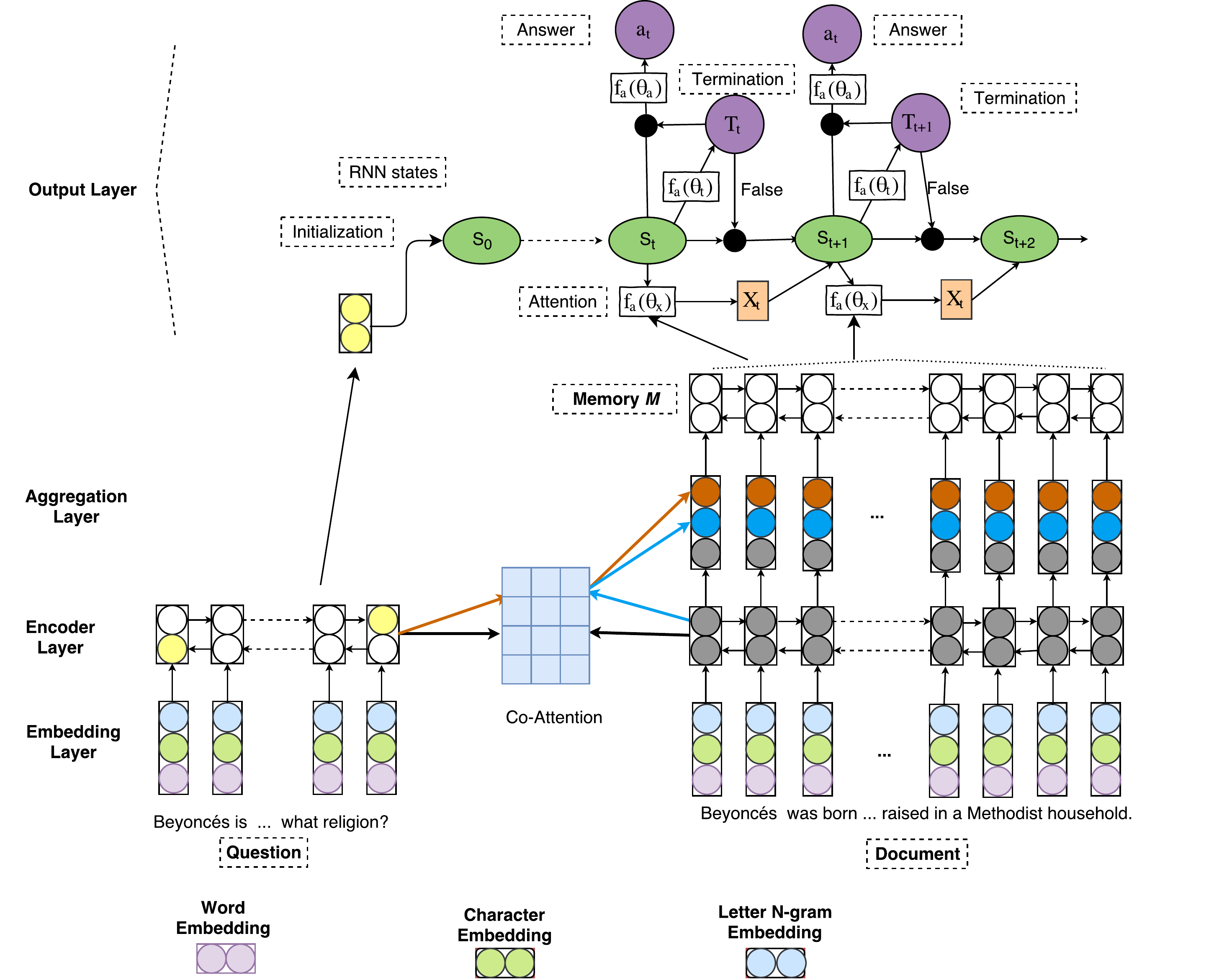}}
\caption{\label{fig:model} {\bf Architecture of \RN}: The embedding/encoder layers compute representations for the question $Q$ and the passage document $P$. The aggregation layer uses co-attention to compute  question-aware passage information and passage-aware question information. Then a GRU combines these information into memory cells and feeds them to the output layer. The output layer models the multiple-turn reasoning mechanism, where intermediate results are stored in $S_t$ and the answer is generated only when the termination signal is triggered. Each $S_t$ is a recurrent network state and models one turn of reasoning.}
\end{figure*}

The 100K (question, passage, answer) tuples is randomly partitioned into a training (80\%), a development (10\%) and test set splits (10\%). Two evaluation metrics are used: Exact Match (EM), which measures the percentage of span predictions that matched any one of the ground truth answer exactly, and Macro-averaged F1 score, which measures the average overlap between the prediction and the ground truth answer. Human performance on the test set is 82.3\% EM and 91.2\% F1.
\\[0.3cm]
\noindent \textbf{MS MARCO}: MS MARCO is a large scale real-world RC dataset that contains 100,000 queries collected from anonymized user logs from the Bing search engine. The characteristic of MS MARCO is that all the questions are real user queries and passages are extracted from real web documents. The data is constructed as follows: for each question/query $Q$, up to approximately 10 passages $P$ are extracted from public web documents and presented to human judges. These passages might potentially have the answer to the question, and are selected through a separate information retrieval system. The judges write down answers in free-form text, and according to the authors of MS MARCO, the complexity of answer varies from a single ``yes/no'' or entity name (e.g. $Q$: ``What is the capital of Italy''; $A$: Rome), to long textual answers (e.g. $Q$: ``What is the agenda for Hollande's state visit to Washington?''). Long textual answers may need to be derived through reasoning across multiple pieces of text.

The dataset is partitioned into a 82,430 training, a 10,047 development, and 9,650 test tuples. Since the answer is free-form text, the evaluation metrics of choice are BLEU \cite{papineni2002bleu} and ROUGE-L \cite{lin2004rouge}. 
To apply the same RC model to both SQuAD (where answers are text spans in $P$) and MS MARCO (where answer are free-form text), we search for spans in MS MARCO's passages that maximizes the ROUGE-L score with the raw free-form answers. Our training data uses these spans as labels, but we evaluate our model with respect to the raw free-form answers; this has an upper bound of $94.23$ BLEU  and $87.53$ ROUGE-L on the dev set. 
By this construction, there are multiple number of passages to read for each question, but the answer span might only involve a few passages (i.e. the ones that include the max ROUGE substring). 
We describe techniques to handle this case in Section \ref{subsec:passage_ranking}.


\section{Model: \RN}
\label{sec:model}

The reading comprehension task involves a question/query $Q=\{{q_0, q_1,..., q_{m-1}}\}$ and a passage $P=\{{p_0, p_1, p_{n-1}}\}$ and aims to find an answer span $A=\left \langle a_{start}, a_{end}\right\rangle$ in $P$. Here, $m$ and $n$ denote the number of tokens in $Q$ and $P$, respectively, while $a_{start}$ and $a_{end}$ indicate the indices of tokens in $P$. The learning process for reading comprehension is to learn a function $f(Q, P)\to A$ trained on a set of tuples $\left\langle Q, P, A\right\rangle$.

Our model \textbf{\RN}, is an extension of ReasoNet \cite{shen2016reasonet} with  three enhancements: (1) In the input layer, we added character and letter 3-gram embeddings to improve robustness to rare words. (2) We implemented co-attention \cite{seo2016bidirectional} in the aggregation layer to focus on relevant words in both $Q$ and $P$. (3) For the MS MARCO task, which needs to handle multiple passages, we incorporated an extra passage ranker component. 
The architecture is shown in Figure~\ref{fig:model}. 
In brief, the embedding/encoder layers first build representations of $Q$ and $P$. The aggregation layer uses co-attention to fuse information from the $Q$-$P$ pair. The output layer is a recurrent net that maintains intermediate state and dynamically decides at which turn to generate the answer. 

\subsection{Detailed description of \RN}

{\bf Embedding Layer:} We adopt three types of embeddings to represent input word tokens in $Q$ and $P$. For word embeddings, we use pre-trained GloVe vectors \cite{pennington2014glove}. To address the out-of-vocabulary problem, we also include character and letter 3-gram embeddings. Character embeddings are fed into a convolutional neural network (CNN) as in \cite{kim2014convolutional}, then max-pooled to form a fixed-size vector for each token. For letter 3-gram embeddings, we follow \newcite{huang2013learning} by first \textit{hashing} each word as a bag of letter 3-gram, then feeding them into another CNN. The concatenation of all embeddings are then fed to a two-layer Highway Network \cite{srivastava2015highway}. Therefore, we obtain the final embedding for the words in $Q$ as a matrix $E^q \in \mathbb{R}^{d \times m}$, and words in $P$ as $E^p \in \mathbb{R}^{d \times n}$, where $d$ is the dimension of the embedding.

\indent{\bf Encoding Layer}: On the top of embedding layer, we utilize a bidirectional Gated Recurrent Unit (GRU), a variety of the Long Short-Term Memory Network (LSTM) {\cite{hochreiter1997long}, to encode the words in context. We obtain $H^q \in \mathbb{R}^{2d\times m}$ as the representation of $Q$ and $H^p \in \mathbb{R}^{2d\times n}$ as the representation of $P$. 

\indent{\bf Aggregation Layer}: In this layer, we construct the memory, a summary of information from both the $Q$ and $P$, for each word in $P$. A co-attention mechanism \cite{seo2016bidirectional}, which attends to $Q$ and $P$ simultaneously, is applied by first computing an alignment matrix in two directions: from $Q$ to $P$ and from $P$ to $Q$. The alignment matrix $C$ measures the similarity between $Q$ and $P$:
\begin{equation}
C = f_{match}(H^q, H^p) \in \mathbb{R}^{m \times n}
\label{eq:align}
\end{equation}
The element at $i$-th row and $j$-th of the alignment matrix, $C_{ij}$, indicates the similarity between $i$-th word in the question and $j$-th word in the passage. In detail, $C_{ij} = f_{match}(H^q_{:i}, H^p_{:j})$ is a trainable scalar function that measures the similarity between two input vectors, $H^q_{:i}$, which is the $i$-th column vector of $H^q$, and $H^p_{:j}$, which is the $j$-th column vector of $H^p$. We parameterize $f_{match}(a, b)=w_C^T[a;b;a \circ b]$, where $\circ$ denotes the Hadmard product, $[;]$ indicates vector concatenation across rows, and $w_C\in \mathbb{R}^{6d}$ is a trainable weight vector.
We normalize $C$ row-wise to produce the attention weight across the passage for each word of the question:
\begin{equation}
C^q = softmax(C) \in \mathbb{R}^{m \times n}.
\label{eq:pqatt}
\end{equation}
To measure which context words in the $P$ have the closest similarity to the words in the $Q$, we define an attention weight on the words in passage as:
\begin{equation}
c^p = softmax(max_{col}(C))^T \in \mathbb{R}^{n}.
\label{eq:pqatt2}
\end{equation}

The final context representation of the $P$ is:
\begin{equation}
\label{eq:eqpass}
U = f_{agg}(H^p, H^qC^q, \sum_{i=0}^{i=m-1}H^p_{:i}c^p) \in \mathbb{R}^{8d \times n}.
\end{equation}
In our experiment, we define {$f_{agg}(B, C, D)=[B ; C ; B \circ C; B \circ {D}]$.
Note that $B, C, D$ are matrices with the same dimension, $\circ$ denotes the Hadmard product and $;$ indicates matrix concatenation across columns.} Note that since {$H^p, H^qC^q, \sum_{i=0}^{i=m-1}H^p_{:i}c^p$} are all $2d$ by $n$ matrices, $U$ is a $8d$ by $n$ matrix. 
Finally, to incorporate the full context, the ``memory cells'' of the passage are computed by a bidirectional GRU on top of $U$:
\begin{equation}
\label{eq:mem}
M^p = BiGRU(U) \in \mathbb{R}^{2d \times n}
\end{equation}

{\bf Output Layer:} This layer dynamically decides when to stop reasoning and output the answer. A recurrent neural network \cite{rumelhart1988learning,elman1990finding} is adopted to maintain the states of the reasoning process. 
Formally, the $t$-th time step of inference state is denoted as $S_t$, and the next state is defined by $S_{t+1} = GRU(S_t, X_t)$. Note that the $X_t$ is an attention vector generated based on the current state and the memory of the passage: $X_t=f_a(S_t, M^p)$ as in \cite{shen2016reasonet}. Specifically, the attention score $a_{t,i}$ on a memory vector $ m_i \in M^p $ given a state $S_t$ is computed as {$a_{t,i} = {softmax}_{i=1,...,|M^p|} \lambda \space $}cosine$({w_1} m_i, {w_2} S_t)$, where $\lambda$ is set to 10 in our experiments and the projection matrices ${w_1}$ and ${w_2}$ map the memory vector and state into the same space, they are learned during training. The attention vector $X_t$ can be written as $X_t = \sum_{i}^{|M|} a_{t,i} m_i$. The initial state $S_0$ of the inference is from the encoding representation of the question (we pick the last state of the forward GRU and the backward GRU in the $H^q$).

The \textit{termination gate} will produce a stochastic random variable according to the current inference state: $T_t \sim p( \cdot | f_t(S_t))$, where $f_t$ is modeled by a $2d \times 10 \times 10 \times 10 \times 1$ feed-forward neural network. Note $T_t$ is a binary random variable: if $T_t$ is true, the recurrent net will stop and the answer model will execute; otherwise it will generate an attention vector $X_{t+1}$ and update the next state $S_{t+1}$.

The answer module needs to output a span in passage. We do this with two feedforward networks, one predicting the start point of the span and the other predicting the end point, so predicted answer at turn $t$ is $a_t=(y^t_{s},y^t_{e})$:
\begin{equation}
\label{eq:start}
y^t_{s} = softmax(w_{s} ^T [M^p, (w_{ps} ^T M^p) \circ S_t])
\end{equation}
\begin{equation}
\label{eq:end}
y^t_{e} = softmax(w_{e} ^T [M^p, (w_{pe} ^T M^p) \circ S_t]).
\end{equation}
where $w_{s}, w_{e}, w_{ps} $ and $ w_{pe}$ are trainable model parameters.
Since the termination state is discrete and is not connected to the final output directly, we use the Contrastive Reward method \cite{shen2016reasonet} inspired by deep reinforcement learning \cite{weissenborn2016separating, mnih2014recurrent} for training.

\subsection{Passage ranking extension}
\label{subsec:passage_ranking}

The MS MARCO dataset provides multiple passages per question/query. Our architecture in Figure \ref{fig:model} is built for a single passage-question pair, so we need to extend it to handle multiple passages. We propose a solution using passage ranking. Assume there are $J$ passages, $P^{(1)},\ldots, P^{(J)}$. First, our model runs independently on every 
$(P^{(j)},Q)_{j=1,\ldots,J}$ pair, generating $J$ different answer spans (empty spans are possible). 
Then, we multiply the probability of each answer span with a score $r(P^{(j)},Q)$ provided by a passage ranker, and output the answer with the maximum combined score, {similar to EpiReader
 \cite{trischler-EtAl:2016:EMNLP2016}}. The passage ranker is a information retrieval model \cite{shen2014latent}.\footnote{Our implementation first hashes words into letter 3-gram (50K dimension), then use a CNN with 256 hidden nodes and the size of window 5, and lastly optimizes the similarity between the vector representations of $P$ and $Q$.} It can be trained on the same RC data, where documents with answers are considered relevant and those that do not are considered irrelevant. 


All our MS MARCO results use the passage ranking extension, unless otherwise mentioned. 

\section{Experiments}
\label{sec:experiment}
\vspace{-0.2cm}
We seek to answer the following questions:
\begin{enumerate}
\item Is multiple-turn reasoning beneficial for RC? (Section \ref{subsec:main_result})
\item What types of questions/answers benefit most from multiple-turn reasoning? (Section \ref{subsec:score_breakdown})
\item How many turns are employed in practice by \RN, and what are the implications for dynamic versus fixed strategies in multiple-turn reasoning? (Section \ref{subsec:distribution})
\end{enumerate}

In addition to the above analyses, we also demonstrate that our \RN~achieves state-of-the-art results (Section \ref{subsec:sota}) and discuss some ablation studies on model variants (Section \ref{subsec:ablation}).\footnote{A note on hyperparameters: Throughout all experiments, we use NLTK to tokenize $P$ and $Q$, 
and employ pre-trained case-sensitive 300 dimension GloVe embeddings\footnote{http://nlp.stanford.edu/data/wordvecs/glove.840B.300d.zip}. A one layer CNN with 100 dimensions and window size of 5 is used to compute the character embeddings; a one layer CNN with 100 dimension and window size of 1 is used for letter 3-gram embeddings. The size of hidden nodes of all GRU's is set to 128. {A five layers feedforward network ($2d(256) \times 10 \times 10 \times 10 \times 1$) } is used for the terminate network and the maximum number of reasoning turns in the recurrent net is capped at 5. To avoid overfitting, we adopt 0.15 dropout rate over the letter 3-gram and character embeddings, and 0.25 dropout rate \cite{srivastava2014dropout} over GRU network. The model is optimized with AdaDelta \cite{zeiler2012adadelta} with an initial learning rate 0.5.}

\begin{table}[t!]
\centering
\begin{tabular}{@{\hskip1pt} l @{\hskip1pt}||@{\hskip1pt} c @{\hskip1pt}|@{\hskip1pt} c @{\hskip1pt}}
\hline
&SQuAD& MS MARCO \\ \hline
Single model	&EM/F1 Score&BLEU/ROUGE-L\\ \hline\hline
Single turn &67.8/76.7& {33.65/36.54}\\ \hline
Fixed 5-turn &70.1/78.9& 34.93/36.67 \\ \hline
\RN  &\textbf{70.8/79.4}& {\textbf{38.62/38.01}} \\ \hline
\end{tabular}
\caption{\label{tab:comp_steps} \textbf{Main results}---Comparison of single turn to multiple turn reasoning strategies on SQuAD and MS MARCO dev sets. Both multiple turn strategies (fixed at 5, or dynamically decided based on \RN) outperform Single turn in all metrics. The dynamic strategy further improves upon the fixed multiple 5-turn strategy.} 
\vspace*{-0.3cm}
\end{table}

\begin{figure}
    \centering
\includegraphics[scale=0.335]{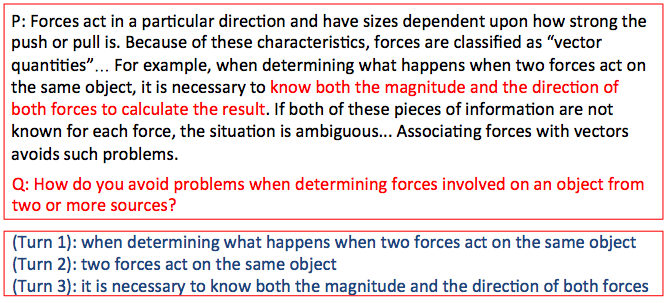}
\caption{\label{fig:casestudy} Case study from SQuAD of answers from multiple turns. In Turn 1, the model identifies a span similar to the question. This is refined and at Turn 3 a better answer becomes attainable.}
\end{figure}

\subsection{Is multiple-turn reasoning beneficial?}
\label{subsec:main_result}

In summary, yes. We compare three systems:\\[0.2cm]
\textbf{Single turn}: the RC model only has one turn of reasoning. This corresponds to a model like Figure \ref{fig:model} without termination nodes, where the output layer always stops at $S_{t=1}$.\\[0.2cm]
 \textbf{Fixed 5-turn}: the RC model runs 5 turns of iterative attention. This is Figure \ref{fig:model} without termination nodes, where output layer always stops at $S_{t=5}$.\\[0.2cm]
 \textbf{\RN~(Dynamic multiple-turn reasoning)}: this is the RC model in Figure \ref{fig:model}, which can decide from 1 to $T$ turns based on the termination probability on each $Q$-$P$ pair at test time. We set $T=5$ to compare with the Fixed 5-turn system. 


The main results are shown in Table \ref{tab:comp_steps}. 
We observe that both multiple turn strategies (either fixed at 5 turns, or dynamically decided based on \RN) outperform the single turn system in all metrics. The dynamic strategy further improves upon the fixed multiple 5-turn strategy. For example, the F1 score on SQuAD improves from 76.7 to 78.9 when increasing the number of turns from 1 to 5, and further improves to 79.4 with dynamic multiple turns. On MS MARCO, we see a ROUGE improvement from 36.54 (1-turn) to 36.67(5-turn) and 38.62 (dynamic multi-turn).
These results convincingly show that multiple-turn reasoning is helpful for SQuAD and MS MARCO tasks. Figure \ref{fig:casestudy} shows a case study of how answers improve with each turn. 

\begin{figure}[t!]
\centering
\begin{subfigure}[b]{0.99\columnwidth}
\includegraphics[scale=0.26]{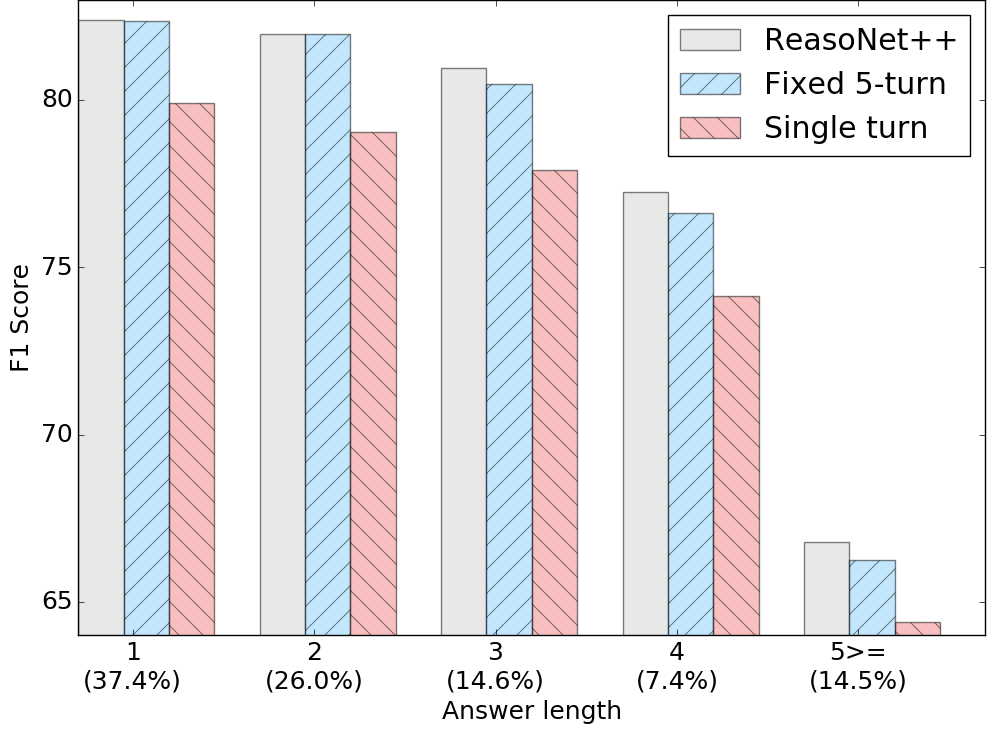}
\caption{SQuAD}
\end{subfigure}
\begin{subfigure}[b]{0.99\columnwidth}
\includegraphics[scale=0.32]{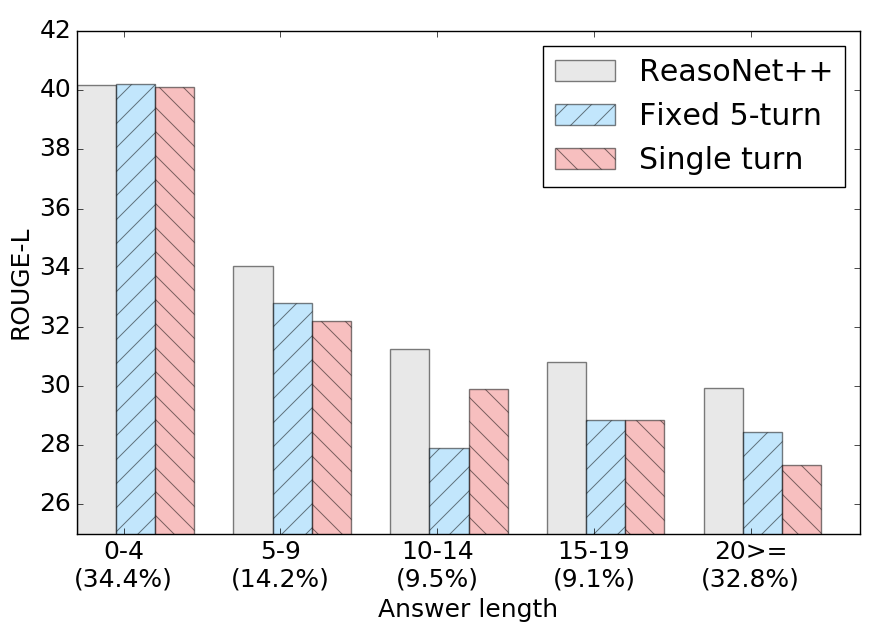}
\caption{MS MARCO}
\end{subfigure}
\caption{\label{fig:score_by_answer_length} Score breakdown by answer length}
\end{figure}



\begin{figure}[h!]
\centering
\begin{subfigure}[b]{0.99\columnwidth}
\includegraphics[scale=0.275]{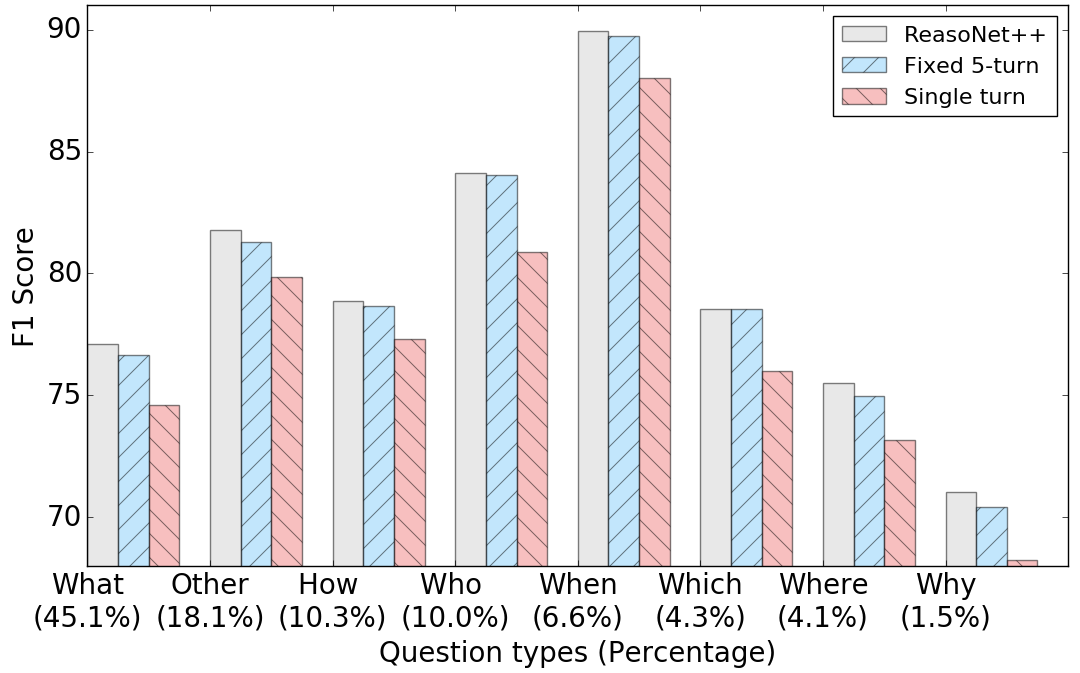}
\caption{SQuAD}
\end{subfigure}
\begin{subfigure}[b]{0.99\columnwidth}
\includegraphics[scale=0.355]{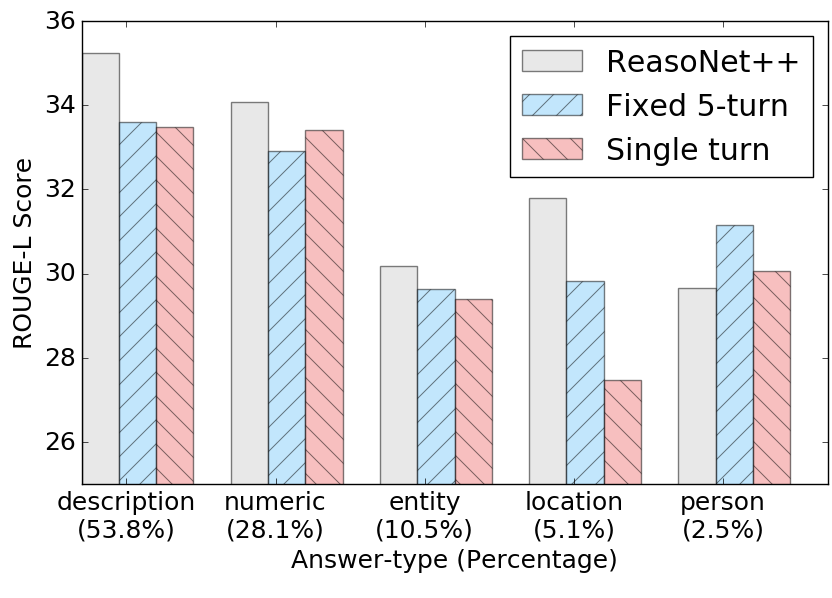}
\caption{MS MARCO}
\end{subfigure}
\caption{\label{fig:score_by_query_type} Score breakdown by query/answer type}
\end{figure}

\subsection{What types of questions/answers benefit most from multiple-turn reasoning?}
\label{subsec:score_breakdown}

We find that improvements from multiple-turn reasoning is generally seen across the board, but particularly helps questions with longer answers.
Figure \ref{fig:score_by_answer_length} shows the score breakdown of Table \ref{tab:comp_steps} according to answer length (\# of words). For SQuAD, both \RN~and Fixed 5-turn outperform Single turn for all answer lengths, and \RN~outperforms Fixed 5-turn for answer lengths $>3$. For MS MARCO,  \RN~outperforms Fixed 5-turn for answer lengths $>5$; on the other hand, there is almost no difference among systems for short answers (0-4). We hypothesize there is a correlation between answer length and the difficulty of the question; for difficult questions there may be more potential for multiple-turn reasoning to improve results. 

We also visualize the score breakdown according to question/answer type (Figure \ref{fig:score_by_query_type}). For MS MARCO, the questions are annotated by the type of the correct answer: description (e.g. Q: ``How to cook a turkey''), numeric (e.g. Q: ``Xbox one release data''), entity, location, person. There is no such annotation for the entire SQuAD dev data, but we can classify questions by their first word: What, Who, When, Which, etc. 
Similar to the answer length results, we observe that multiple-turn reasoning outperforms single turn for SQuAD across the board, regardless of question type. For MS MARCO, \RN~ gave large improvements over single turn in particular for description and location types. {Descriptions tend to be lengthy, so this again corroborates our hypothesis that there may be more potential gains for questions requiring long answers.}

\subsection{How many turns of reasoning are employed in practice?}
\label{subsec:distribution}

We are interested in understanding the number of turns determined by \RN. When does it decide to terminate? In Figure \ref{fig:step_dist}, we plot the distribution of turns until the model decides on start points and end points (of the answer span).

First, note the start point is often decided before the end point, e.g. the start is already determined at turn 3 for approximately 20\% of the questions , but the end does not get predicted until turn 4 or 5. Intuitively, we think it is easier to first identify the start of an answer, then use that signal as intermediate state $S_t$ to identify the end point.

Second, there is almost no termination at turns 1 or 2, implying the model prefers more iterations of reasoning. Most terminations are done at step 4 or 5, which explains the relatively close performance results between Fixed 5-turn and \RN.

\begin{figure}[t!]
    \centering
\includegraphics[scale=0.34]{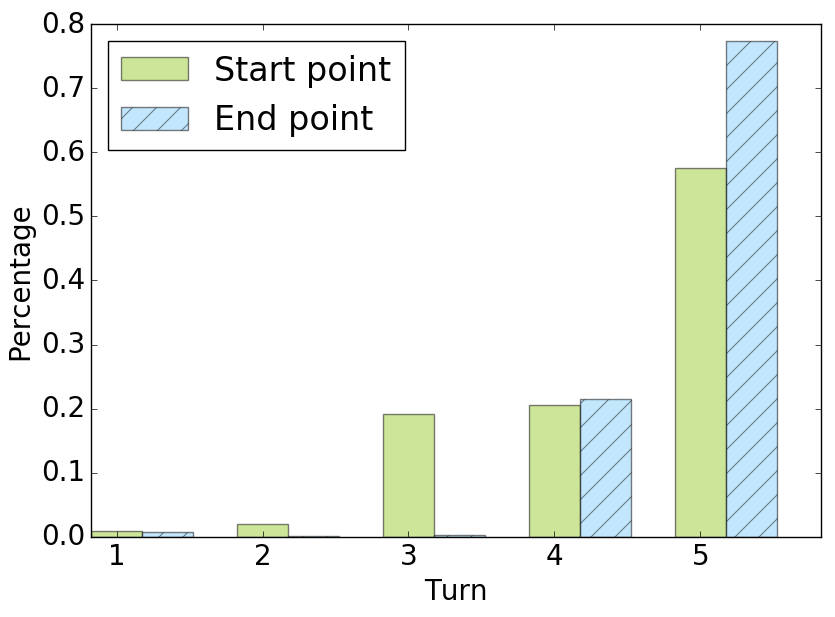}
\caption{\label{fig:step_dist} Distribution on the number of turns by \RN on the SQuAD dev set. Note that start points are often decided before end points, and most answer spans are generated after 3 turns.}
\end{figure}

\subsection{Comparison with state-of-the-art}
\label{subsec:sota}

Our \RN~model, which is an extension of ReasoNet \cite{shen2016reasonet}, achieves scores competitive with state-of-the-art results. The official leaderboard results are shown in Table \ref{tab:marco_result} (MS MARCO) and Table \ref{tab:squad_result} (SQuAD)
Results are divided by whether we use an individual model or an ensemble of models. 
For SQuAD, the \RN~ensemble model achieves the best EM and F1 test score among all published works, and places second if we include r-net. Similarly, the \RN~individual model results are in the top 1 or 2 ranks, competitive with published works like \newcite{zhang2017exploring} and \newcite{weissenborn2017fastqa}. 
{For MS MARCO (Table \ref{tab:marco_result}), \RN~ranks first in test ROUGE and second in test BLEU (after Match-LSTM \cite{wang2016machine}).}
Note that some of the models on the leaderboard use multiple-turn reasoning, while others do not. But we refrain from drawing conclusions about multiple-turn reasoning by comparing across models, due to other confounding variables, e.g. different embeddings and network architectures. 


\begin{table}[t!]
\centering
\begin{tabular}{@{\hskip1pt} l @{\hskip1pt}||@{\hskip1pt} c @{\hskip1pt}|@{\hskip1pt} c  }
\hline
\multirow{2}{*}{System}&\multicolumn{2}{c}{BLEU/ROUGE-L} \\ \cline{2-3}
           	&Dev Set& Test Set\\ \hline\hline
{\textbf{\RN~$^{Individual}$}} &38.62/38.01&39.86/38.81 \\ \hline
         	
Match-LSTM& -/-&40.72/37.33\\ \hline
FastQA\_Ext &35.0/34.4 &33.93/33.67 \\ \hline
FastQA& 34.9/33.0&33.99/32.09 \\ \hline
\hline
Human Performance &-/-&46/47\\ \hline
\end{tabular}
\caption{\label{tab:marco_result} Official MS MARCO leaderboard performance on April 5, 2017.}
\vspace*{-0.3cm}
\end{table}

\begin{table*}[t!]
\centering
\begin{tabular}{l || c | c  }
\hline
Ensemble model results:&Dev Set (EM/F1)& Test Set (EM/F1) \\ \hline
\hline
r-net* & -/- & 76.9/84.0 \\ \hline
\textbf{\RN~(Ensemble model)} & 75.4/82.9&75.0/82.6 \\ \hline
BiDAF \cite{seo2016bidirectional}&73.3/81.1 &73.7/81.5 \\ \hline
Multi-Perspective Matching \cite{wang2016multi} &69.4/78.6 &73.8/81.3 \\ \hline
Dynamic Coattention Networks \cite{xiong2016dynamic} & 70.3/79.4 &71.6/80.4\\ \hline
Match-LSTM with Ans-Ptr \cite{wang2016machine} &67.6/76.8&67.9/77.0\\ \hline
Fine-Grained Gating\cite{yang2017differentiable} &62.4/73.4 &62.4/73.3\\ \hline \hline
\multicolumn{3}{l}{\textit{Individual model results:}} \\ \cline{1-3}
r-net* & -/-&72.3/80.7 \\ \hline
jNet \cite{zhang2017exploring}&-/- &70.6/79.8\\ \hline
Ruminate Reader* & -/-&70.6/79.5 \\ \hline
\textbf{\RN~(Individual model)} &70.8/79.4 &70.6/79.36\\ \hline
Document Reader*  &-/-&70.7/79.35 \\ \hline
FastQAExt \cite{weissenborn2017fastqa} &70.3/78.5&70.8/78.9 \\ \hline
RaSoR \cite{lee2016learning} &66.4/74.9&70.0/77.7 \\ \hline
BiDAF \cite{seo2016bidirectional} &67.7/77.3&68.0/77.3\\ \hline
Iterative Co-attention Network* &-/-&67.5/76.8 \\ \hline
Dynamic Coattention Networks \cite{xiong2016dynamic}& 65.4/75.6 &66.2/75.9 \\ \hline
Match-LSTM with Bi-Ans-Ptr \cite{wang2016machine} &64.1/73.9&64.7/73.7 \\\hline
Attentive CNN context with LSTM* &-/-& 63.3/73.5 \\ \hline
Dynamic Chunk Reader \cite{wang2016machine} &62.5/71.2&62.5/71.0 \\ \hline
LR baseline \cite{rajpurkar2016squad} &40.0/51.0&40.4/51.0 \\ \hline
 \hline
Human Performance &80.3/90.5&82.3/91.2\\ \hline
\end{tabular}
\caption{\label{tab:squad_result} Official SQuAD leaderboard performance on April 5, 2017. Asterisk * denotes unpublished works. Results are sorted by Test F1.}
\vspace*{-\baselineskip}
\end{table*}

\subsection{Ablation studies and model variants}
\label{subsec:ablation}

We now present some ablation studies to demonstrate the differences between our \RN~and the original ReasoNet \cite{shen2016reasonet} in which we are based on.\footnote{Due to time constraints, we only perform ablation studies on the embedding and passage ranking enhancements, and leave the study of the impact of co-attention to future work.} 

First, Table \ref{tab:comp_embed} shows the improvement from adding sub-word level modeling to ReasoNet, which only used word embeddings. We observe marked improvements of {update number ~+1.1 F1 in SQuAD and +0.9 ROUGE in MS MARCO}. Although these improvements are not as large as those we achieved with multiple-turn reasoning, they are are still considerable and imply that robust representations of words is an important building block to strong RC models. 

\begin{table}[t!]
\centering
\begin{tabular}{@{\hskip1pt} l @{\hskip1pt}||@{\hskip1pt} c @{\hskip1pt}|@{\hskip1pt} c @{\hskip1pt}}
\hline
\multirow{2}{*}{System}&SQuAD& MS MARCO \\\cline{2-3}
	&EM/F1 Score&BLEU/ROUGE\\ \hline
word+char+3gram  &70.8/79.4& {38.62/38.01}\\ \hline
word+char &70.4/79.1&{38.37/37.91}\\ \hline
word &69.9/78.3&{37.77/37.14} \\ \hline
\end{tabular}
\caption{\label{tab:comp_embed} Comparison of input embeddings: the addition of character (char) and letter trigram (3gram) embeddings to word embeddings (word) clearly improve results on SQuAD and MS MARCO development sets. } 
\vspace*{-0.3cm}
\end{table}

Secondly, Table \ref{tab:marco_exp} shows the impact of passage ranking---this is only relevant for MS MARCO, which contains multiple passages for each question/query. 
Recall that the RC model needs to read approximately 10 passages to answer each query, and on average only one or two passage contain answer spans. 
\RN~extracts answer spans from each passage independently, then combines with an IR model to output the final answer. If we assume oracle ranking from the IR model, we can achieve 62 BLEU / 63 ROUGE, suggesting that better passage ranking models (e.g. via joint training with RC models) is fruitful as future work.

\begin{table}[t!]
\centering
\begin{tabular}{l | c }
\hline
	& BLEU/ROUGE-L\\ \hline
Oracle passage selection & 62.83/63.17\\ \hline 
Passage ranking &{38.62/38.01} \\ \hline
\end{tabular}
\caption{\label{tab:marco_exp} Effect of multiple passages per query in MS MARCO. } 
\vspace*{-0.5cm}
\end{table}

\section{Conclusion}
\label{sec:conclusion}

This paper empirically investigates the performance of single-turn and multiple-turn reasoning on two challenging reading comprehension tasks: SQuAD and MS MARCO. 
To perform the analysis, we adopt the neural network model of \newcite{shen2016reasonet}, 
which employs iterative attention and uses reinforcement learning to dynamically control the number of
turns. We find that multiple-turn reasoning outperforms single-turn reasoning for all question and answer types; further, we observe that enabling a flexible number of turns generally improves upon a fixed multiple-turn strategy. 
While our analysis is based on a single model, we believe the conclusions will be valuable for most RC methods using attention-based neural network. 
Our model extension to \cite{shen2016reasonet} achieves results competitive to the state-of-the-art on both tasks. 
As future work, we plan to investigate the impact of even deeper layers of reasoning and explore fast training methods to make such methods practical for large-scale datasets.


\bibliography{re-emnlp2017}
\bibliographystyle{ijcnlp2017}

\end{document}